# Epistemic AI platform accelerates innovation by connecting biomedical knowledge

*Da Chen Emily Koo, Heather Bowling, Kenneth Ashworth, David J. Heeger, Stefano Pacifico*

*Epistemic AI*

## Abstract

Epistemic AI accelerates biomedical discovery by finding hidden connections in the network of biomedical knowledge. The Epistemic AI web-based software platform embodies the concept of *knowledge mapping*, an interactive process that relies on a knowledge graph in combination with natural language processing (NLP), information retrieval, relevance feedback, and network analysis. Knowledge mapping reduces information overload, prevents costly mistakes, and minimizes missed opportunities in the research process. The platform combines state-of-the-art methods for information extraction with machine learning, artificial intelligence and network analysis. Starting from a single biological entity, such as a gene or disease, users may: a) construct a map of connections to that entity, b) map an entire domain of interest, and c) gain insight into large biological networks of knowledge. Knowledge maps provide clarity and organization, simplifying the day-to-day research processes.

## Introduction

Biomedical knowledge is the key to accelerate discoveries and identify hypotheses leading to novel diagnostic tools, therapies, and vaccines. Biomedical researchers struggle with information overload while attempting to grapple with the vast and rapidly expanding base of biomedical knowledge including documents (e.g., papers, patents, clinical trials) and databases (e.g., databases of genes, proteins, pathways, drugs, diseases, medical terms). This is a major pain point for biomedical researchers and, with no appropriate solution available, they are forced to use search tools (e.g., PubMed and Google Scholar) and cumbersomely navigate manually-curated databases. These tools are suitable for finding documents matching keywords (e.g., a single gene or published journal paper), but not for acquiring a collection of knowledge to explore and learn about a topic area or subdomain (e.g., the long term impact of COVID-19 on cognitive function), or for interpreting the results of high-throughput biology experiments such as gene sequencing, protein expression, or compound screening.

We developed the Epistemic AI platform to remedy this problem. The platform is an innovative, AI-powered and interactive platform for researchers to connect with and discover knowledge more quickly and efficiently than ever before, saving time and cost, while also providing completeness and accuracy to make informed decisions, especially at critical stage-gates. This platform is changing the way that biomedical investigators work and think, thereby fueling the next generation of breakthrough discoveries and innovations to improve human health.

The core innovation in the Epistemic AI platforms is what we call *knowledge mapping*.

Knowledge mapping uses a knowledge graph in combination with biomedical natural language processing (bioNLP) (Devlin et al, 2018; Przbyła et al, 2018; Afentenos et al, 2005; Elhadad et al, 2005; Yetisgen-Yildiz and Pratt, 2005; Goldstein, 2007; Roberts et al, 2007; Kipper-Schuler et al, 2008; Fiszman et al, 2009; Savova et al, 2010; Luther et al, 2011; see Bretonnel Cohen and Demner-Fushman, 2014 for review), relevance feedback (Agichtein, Brill and Dumais, 2006; States et al, 2009; Yu et al, 2009; Alatrash et al, 2012; Ji et al, 2016; Rocchio et al, 1971), and network analysis (Zhong et al, 2006; Mostafavi et al, 2008; Suthram et al, 2010; Cokol et al, 2011; Green et al, 2011; Ciofani et al, 2012; Marbach et al, 2012; Guimera And Sales-Pardo, 2013; Kurts et al, 2015; Shi et al, 2016; Suresh et al, 2016; Wong et al, 2016; Wang et al, 2017; Gligorijević et al, 2018; Castro et al, 2019; Chasman et al, 2019; Cramer et al, 2019; Miraldi, 2019; Siahpirani et al, 2019), for an interactive knowledge mapping platform.

## Methods

**Technical challenges**

We faced four critical technical challenges in developing the platform:

1) Biomedical knowledge is siloed, locked in text documents or spread across multiple biomedical databases that each provide partially overlapping and complementary information. We addressed this problem by offering a seamless integration of knowledge extracted from text documents and a multitude of structured biomedical databases with a hybrid architecture combining deep learning and statistical machine learning (ML) with knowledge-intensive AI methods (Gabor et al, 2018; Luan et al, 2018).

2) Despite the advances in natural language processing (NLP) and bioNLP (see Introduction for references), a fully automatic solution with acceptable performance for knowledge extraction, representation, and reasoning is not close to being realized (Weischedel and Boschee, 2018), e.g., the failure of IBM Watson Health (Strickland, 2019). Current methods perform best when trained on large, manually-labeled datasets but labeling is time-consuming, especially in highly specialized tasks such as those in BioNLP. Precision and recall of BioNLP is inadequate (Doğan et al, 2019; Wang et al, 2019), and automated reasoning suffers from inadequately modeling the complexity of both the data and the problem domain (Rocktäschel and Riedel, 2017) – the "blocks world" problem of symbolic AI (Slaney and Thiébaux, 2001). To circumvent these problems, rather than fully automating knowledge discovery, our approach relies on the premise that machine amplification of skilled experts can unlock enormous value. Consequently, we utilize an interactive platform; precision and recall are increased as the investigator adds *landmarks* to their knowledge map, and the human domain expert does the hard reasoning. Rather than attempt to replace the expertise of biomedical researchers with a fully automated system, the Epistemic AI platform augments the user's expertise with an interactive process for knowledge mapping and knowledge discovery. This lightens the demands of the AI/ML technology, so that it is fully capable of supporting the platform, with a human-in-the-loop.

3) Information retrieval and ranking is context- and task-dependent – two users submitting the same query may have different goals, and consequently are seeking different information. Our knowledge mapping platform deals with this naturally via user interaction. An investigator begins by entering terms in a text box. Our information retrieval algorithms find relevant biomedical entities, documents and relationships. The investigator may then select any of the results and add them as landmarks in their knowledge map, providing context to re-rank the search results. This process is repeated iteratively with the option to replenish (re-rank) the results with respect to proximity to the landmarks in the map at each stage. Precision and recall (measures of relevancy) increase through the interaction as the investigator adds more landmarks to the map, making it easier to identify and consolidate all of the relevant knowledge. This process is similar to "relevance feedback", which combines search with explicit supervision from users to indicate relevant or useful results (Rocchio, 1971; Agichtein, Brill and Dumais, 2006; States et al, 2009; Yu et al, 2009; Alatrash et al, 2012; Ji et al, 2016). Explicit relevance feedback was not widely adopted in traditional text search applications (Spink, Jansen and Ozmultu, 2000; Anick, 2003), but we are tackling a different problem (knowledge mapping, not search) that inherently involves user interaction and exploration in a manner that is more organic than previous attempts with text search. Critically, the end user (an expert biomedical investigator) is the ultimate arbiter of quality and accuracy, deciding at each step whether to include a particular entity or relationship in their map.

4) Rapid progress in biomedical R&D requires an interdisciplinary approach but knowledge and expertise are siloed. For example, the domain of relevant knowledge for COVID-19 is vast and interdisciplinary, including virology, immunology, pulmonology, molecular biology, and so on. In spite of incredible progress with vaccines, we only have a limited understanding of disease progression and its long-term consequences. Only a minority of investigators have the breadth of knowledge and interdisciplinary mindset to generate hypotheses about the consequences and possible treatment for long-COVID (i.e., the effects of COVID-19 infections that continue for weeks or months beyond the initial illness). And those highly interdisciplinary investigators may lack the depth to see it through. Researchers are often required to work with incomplete, ambiguous, anomalous or sometimes even deceptive data. Moreover, as with the current environment with COVID-19, time constraints require quick decisions that can conspire with natural human biases, leading to inaccurate or incomplete judgements. The Epistemic AI platform enables researchers to overcome these problems by helping them to identify and refute as many possible hypotheses as possible given a full range of data, information, assumptions and gaps pertinent to the problem at hand. The platform enables investigators to obtain a complete map of the relevant literature and knowledge, including links to related fields and topics. Such an interdisciplinary viewpoint is a key driver to discovery, increasing the likelihood of gaining critical insights.

**Solutions**

Our scientific knowledge mapping platform addresses these challenges with four technical innovations. Although the technology underlying each of these four innovations has a long history of development, their combined usage represents a major advancement on the state-of-the-art. All together, this aggregate, combining all four, is transformative.

1) *Knowledge graphs and knowledge mapping*. Knowledge graphs are not new, and the concept of knowledge mapping is not new (investigators map knowledge in their brains and/or with the help of spreadsheets or reference managers every second of the day), but our novel approach that uses a knowledge graph in combination with BioNLP, relevance feedback, and network analysis, for an interactive knowledge mapping platform, has not been previously developed and commercialized. Knowledge mapping is quite different from conventional search, and solves a critical problem for users. For example, the conventional process of literature search becomes exponentially cumbersome with more documents in a collection because there are more leads to chase down. The investigator has to look through all the references in all the documents in their collection, identify those that they have not yet read, choose a subset to read, and then repeat the process ad infinitum. With knowledge mapping, this process is radically simplified. Relevant knowledge is tightly connected (in close proximity) to entities in a knowledge map, so identifying additional relevant knowledge becomes easier (more robust and reliable) as the investigator adds entities to their map. In addition, the Epistemic AI knowledge graph contains information ingested from myriad databases in addition to documents, so a biomedical researcher may draw on all of these sources seamlessly during the knowledge mapping process. Provenance (a link to the original source) is always provided.

2) *Information extraction*. The Epistemic AI platform uses state-of-the-art and novel information extraction algorithms to create a network of biomedical knowledge by extracting entities and relations from documents and databases. Though information extraction from text is far from perfect, users of the Epistemic AI platform have validated that the technology is a huge improvement compared to previous workflows. As the technology improves over time, its application in this interactive platform will scale accordingly.

3) *Proximity ranking*. We have developed a novel machine learning algorithm for ranking documents and biomedical entities by calculating conceptual proximity from language and network features. This ranking algorithm is embedded in each step of an iterative process with a human-in-the-loop.

## Results

We developed a human-in-the-loop responsive research platform that allows the user to specify the most important biomedical entities and parameters and to examine connections between biomedical entities.

To illustrate how the platform works, we considered the question "Are Alzheimer's disease and dementia patients more at risk for poor outcomes with COVID-19?" We constructed a knowledge map (KM) by adding "COVID-19", "Alzheimer's Disease", and "Dementia" as Disease entities and began by exploring publications and clinical trials (Fig 1A). Specific publications related to dementia and COVID-19 were surfaced by the Epistemic AI platform and an interactive feature permitted us to select publications we deemed most relevant to our query. This was achieved simply by clicking the star next to the publication of interest to add it to the KM (Fig 1B-C). The platform then also retrieved other publications and clinical trials that were similar to our publication of interest (e.g., containing closely related entities) and re-ranked the results according to their relevance to the entire KM (Fig 1D). As more publications and clinical trials were added to the KM, the results became increasingly refined, minimizing the need to scroll through irrelevant information and read non-related abstracts.

Given the wealth of publications and clinical trials surfaced by the Epistemic AI platform, we also wanted to examine the relationships between entities within the KM more closely. To achieve this, we performed a deep dive to identify and add other associated entities to the KM. For instance, in this use case, the platform displayed a columnar "card" for each disease entity: COVID-19 (Fig 2A), Alzheimer's Disease, and Dementia. Each of these cards contained various sections with information specific to each individual disease but re-ranked according to the context of the entire KM. For example, the first related publication in the COVID-19 card concerned both COVID-19 and Dementia (Fig 2A). In addition, researchers also have access to, but not limited to, related clinical trials as well as associated genes (Fig 2B) and drugs, any of which can be seamlessly added to the existing KM by clicking its star to explore further connections.

In this use case, we starred one associated gene, IL-6, and presented an example of a gene card with its summary and related variants (Fig 2C). Because all the associated entities presented on the platform are ranked by relevance to the KM, the most relevant associations also appear at the top of each section of each card (Fig 2D).

Using our platform, we determined which common bad outcome risk factors are highest for patients with dementia and COVID-19: CRP and IL-6. We then explored how IL-6 interacts with the immune system and discovered that it shares a common immunological response pathway with CRP. In addition, the platform revealed a new research opportunity: there is an open question of whether COVID-19 infection increases the risk of dementia but a lack of sufficient long-term data to reach a conclusion.

The example provided here illustrates how any biomedical researcher can easily create a knowledge map starting with a simple query about the increased severity of COVID-19 for dementia patients. The Epistemic AI platform rapidly unearthed relevant publications, clinical trials, molecular connections, potential biomarkers, and interventions, alleviating the need to navigate between multiple web pages. Researchers can then not only explore existing connections effortlessly, but also identify new opportunities with unanswered research questions in the field.

With the overwhelming body of research into COVID-19 and dementia spanning publications, clinical trials, and potential molecular mediators, the average researcher would have required hours, if not days, to weed through multiple sources to retrieve all the relevant knowledge on this topic using conventional methods. The Epistemic AI platform, by contrast, enables the user to acquire the same knowledge within minutes.

The Epistemic AI platform automatically identifies relevant connections in the network of biomedical knowledge, so that the user does not have to stochastically locate them across multiple platforms or read dozens of papers to surface the same information. The Epistemic AI platform guides users to relevant information and connected entities, allowing them to seamlessly transition between different types of information, ultimately saving time and mental energy.

**Figure 1. Exploring Publications and Clinical trials within the Epistemic AI platform.** *We entered COVID-19, Alzheimer's Disease and Dementia as Disease entities into the knowledge map (KM) (A). Upon querying, the platform surfaced relevant publications which could be starred and added to the KM by the user (B). Once the user-selected publications were added to the KM (C), the Epistemic AI platform identified and re-ranked the results to provide even more relevant publications and clinical trials (D).*

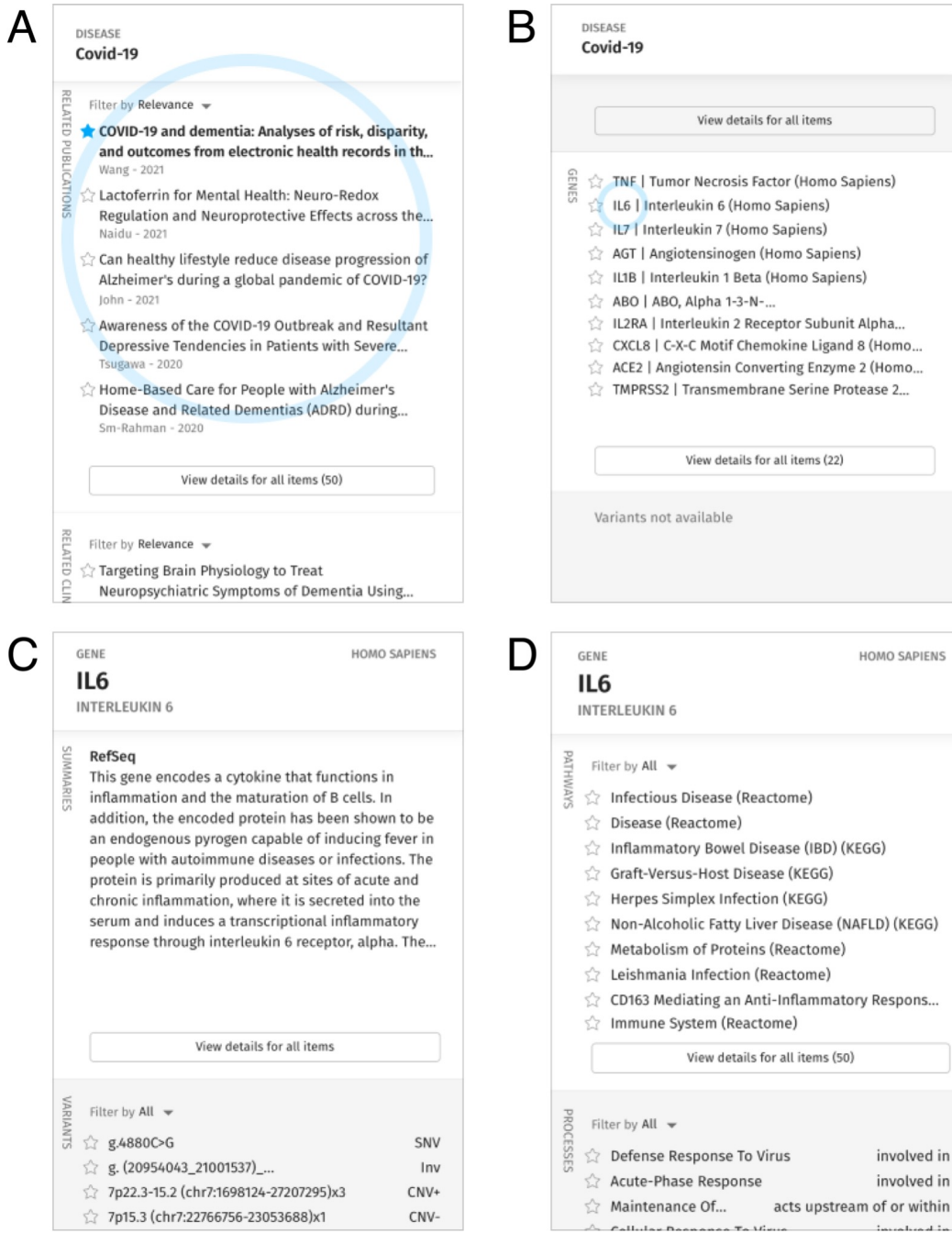

**Figure 2. Performing a "deep dive" analysis on the Epistemic AI platform between entities in the KM.** *Every entity (e.g., COVID-19 disease) added to the KM has a columnar card containing various sections with information specific to that entity (e.g., related publications), but in the context of the rest of the KM (A). Entities within any section (e.g., IL-6 gene), can be added to the KM for further exploration (B). Once added, a new card for IL-6 is created. Example sections include summaries of its protein function and related variants (C), as well as related pathways and processes (D).*

## Conclusions

The Epistemic AI platform has the potential to revolutionize the process by which biomedical research data is aggregated, utilized, analyzed, and presented. We can lower the barrier to insights by reducing the mental load of researchers trying to find relevant information and connect the dots between biomedical entities.